\documentclass[a4paper,fleqn]{cas-dc}
\usepackage[numbers]{natbib}
\usepackage{array}
\usepackage{float}
\usepackage{graphicx}
\usepackage{longtable}  
\usepackage{amssymb}
\usepackage{subcaption} 
\usepackage{xcolor}  
\usepackage{colortbl}  
\usepackage{pifont} 
\usepackage{lipsum}  
\usepackage{amsmath, algorithm, algpseudocode}
\usepackage{multirow}               
\usepackage{etoolbox}  
\usepackage{makecell}

\def\tsc#1{\csdef{#1}{\textsc{\lowercase{#1}}\xspace}}
\tsc{WGM}
\tsc{QE}
\tsc{EP}
\tsc{PMS}
\tsc{BEC}
\tsc{DE}
\usepackage{tabularx}
\newcolumntype{L}[1]{>{\raggedright\arraybackslash}p{#1}}

\begin{document}

\shorttitle{VectorRAG and GraphRAG based Modeling}

\shortauthors{Islam and Sarker et~al.}

\title [mode = title]{Enhancing LLMs with Context-Specific Knowledge for Mitigating Misinformation in SMEs: A RAG-based Modeling and Analysis}



%

\author[1]{Md. Samiul Islam}[]
\author[1]{Iqbal H. Sarker}[
                        orcid=0000-0003-1740-5517]
\cormark[1]
\ead{m.sarker@ecu.edu.au}

\author[1]{Chadni Islam}[]
\author[1]{Ahmad Mohsin}[]
\author[1]{Ahmed Ibrahim}[]
\author[1]{Helge Janicke}[]



\affiliation[1]{organization={School of Science, Edith Cowan University},
    city={Perth},
    postcode={WA-6027}, 
    country={Australia}}

\cortext[cor1]{Corresponding author}

\begin{abstract}
Large Language Models (LLMs), a part of artificial intelligence (AI), are increasingly being adopted by Small and Medium Enterprises (SMEs) to enhance question-answering capabilities and support business decision-making processes. However, hallucinations in LLM-generated outputs can serve as a source of misinformation, reducing user confidence in their reliability and trustworthiness within SMEs. Retrieval-Augmented Generation (RAG) has emerged as a promising approach to address this challenge by incorporating external knowledge sources into the modeling process. In this paper, we present VectorRAG and GraphRAG modeling approaches to mitigate hallucinations and misinformation risks and evaluate their effectiveness in SME environments. Our experimental evaluation is conducted on multiple state-of-the-art LLMs, including LLaMA, Mistral, and Qwen, to assess performance in terms of useful response generation, risk of hallucination, contextual relevance, as well as human-interpretation. The results demonstrate that RAG-enhanced LLMs can significantly improve response quality by reducing hallucinations and misinformation, thereby supporting more reliable, trustworthy, and context-aware decision-making in SME environments.
\end{abstract}


\begin{keywords}
Misinformation \sep Hallucination \sep Artificial Intelligence (AI) \sep Large language models (LLMs) \sep Retrieval-Augmented Generation (RAG) \sep GraphRAG \sep Cybersecurity \sep Business Analytics \sep Small and Medium Enterprises (SMEs)
\end{keywords}

\maketitle

\section{Introduction} 
SME is generally defined as a business organization with a relatively small workforce, limited revenue, and moderate operational scale compared to large enterprises. Representing approximately 98\% of all registered businesses in Australia, this SME sector plays a disproportionately significant role in driving economic activity, innovation, and employment, while contributing substantially to national gross domestic product (GDP) ~\cite{c101}. While SMEs are fundamental to economic growth and innovation, they frequently lack the resources and specialized knowledge required to effectively leverage emerging technologies while keeping pace with rapidly changing digital environments ~\cite{c102}.

In recent years, LLMs have emerged as a promising technology for SMEs, supporting a wide range of tasks, including automated customer interactions, content generation, business decision-making, and knowledge-based assistance across diverse operational domains ~\cite{c2} ~\cite{c103}. Models such as LLaMA, Mistral, and Qwen generate human-like texts that made them the cost-effective and scalable tools to enhance productivity ~\cite{c103}. Despite their remarkable capabilities, LLMs remain challenged by their reliance on large pre-trained datasets, which may not adequately capture the domain-specific knowledge and contextual nuances of SME environments. For instance, an LLM may produce misleading information such as outdated cybersecurity recommendations, misinterpret industry-specific regulations, generate inaccurate compliance advice, or fail to consider SME-specific policies and operational contexts, leading to potentially inappropriate responses.

A particular concern is hallucination, where LLM models may unintentionally generate inaccurate, fabricated, or misleading information based on the data on which they were trained, that undermine trust and facilitate the dissemination of misinformation ~\cite{sarker2024llm}~\cite{c104}~\cite{c9}. This can occur without malicious intent and result in the spread of false or harmful information, further exacerbating the risks associated with these models ~\cite{c104}~\cite{c17}. These misleading outputs can adversely affect decision-making processes and pose challenges to the safe, trustworthy, and responsible use of LLMs~\cite{sarker2024llm}~\cite{c106}.

Retrieval-Augmented Generation (RAG) has emerged as a promising solution that enhances LLMs through the incorporation of external knowledge sources \cite{uddin2026fine}. By combining the model’s internal parametric knowledge with retrieved non-parametric information, RAG enables the generation of more accurate, reliable, and contextually relevant responses. Among RAG techniques, vector approaches rely on semantic similarity search to retrieve relevant information from unstructured data, while graph approaches utilize knowledge graphs to enable structured reasoning over relationships between entities. Despite their advantages, both approaches exhibit inherent limitations. Vector-based methods often struggle to capture explicit semantic relationships among entities, whereas graph-based methods may encounter scalability challenges and limitations arising from sparse or incomplete relational structures. Consequently, ensuring scalable yet contextually accurate retrieval remains a critical challenge for deploying responsible and trustworthy LLM-based solutions in SME environments.

A structured and systematic framework is therefore necessary to facilitate the secure and responsible deployment of LLMs in SMEs. In our previous work, Sarker et al. \cite{sarker2026sme} introduced the SME-TEAM framework, a multi-phased approach founded on four key pillars - Data, Algorithms, Human Oversight, and Model Architecture - to leverage trust and ethics for the secure and responsible use of AI and LLMs in SMEs. Building upon this foundation, the present study conducts a systematic experimental evaluation of VectorRAG and GraphRAG approaches to assess their effectiveness in mitigating misinformation, reducing hallucinations, and enhancing contextual relevance in SME applications. The findings provide practical insights into the design of trustworthy RAG-enhanced LLM systems and contribute to the secure and responsible adoption of AI technologies in SME environments.

Overall, this study makes the following key contributions:

\begin{itemize}
    \item \textbf{RAG-Based Modeling:} We develop and evaluate both VectorRAG and GraphRAG approaches, providing insights into their effectiveness within SME-specific application contexts.

    \item \textbf{Hallucination and Misinformation Mitigation:} We investigate the use of RAG-enhanced LLMs to improve factual grounding, reduce hallucinations, and mitigate misinformation in SME environments.

    \item \textbf{SME-Centric Experimental Evaluation:} We conduct extensive experiments using SME-focused datasets to benchmark and compare the performance of state-of-the-art LLMs, including LLaMA, Mistral-7B, and Qwen.
\end{itemize}

The remainder of the paper is organized as follows. Section 2 presents the literature review. Section 3 describes the methodology adopted in this study. Section 4 reports the experimental results, while Section 5 discusses the key insights derived from the findings. Finally, Section 6 concludes the study.

\section{Related Work}

LLMs have significantly improved natural language understanding and generation capabilities. But there is still some lacking about misinformation and hallucination. Recent studies have increasingly explored RAG, knowledge-enhanced reasoning, hallucination mitigation frameworks, and misinformation detection strategies to improve the factual grounding and trustworthiness of LLM-generated responses.

Wan et al. (2025) \cite{c7}  proposed a hybrid RAG framework that combines vector retrieval with knowledge graphs to improve contextual reasoning and question-answering performance in smart manufacturing environments. Their framework demonstrated that integrating structured and unstructured knowledge sources can significantly improve retrieval precision and contextual relevance in domain-specific tasks. Similarly, Upadhyay and Viviani (2025) \cite{c13} introduced a RAG-based Health Information Retrieval (HIR) system that prioritizes both topical relevance and factual accuracy through semantic similarity and stance-aware ranking mechanisms. Their findings highlight the importance of retrieval-grounded generation for reducing misinformation while improving explainability in sensitive domains such as healthcare. These studies collectively demonstrate that external knowledge integration plays a critical role in improving factual consistency and domain adaptation in LLM systems.

Zhang and Zhang (2025) \cite{c7} presented a comprehensive review of hallucination issues in retrieval-augmented LLMs, identifying retrieval quality, prompt ambiguity, and generation-stage inconsistencies as key contributors to hallucinated responses. Their work emphasizes prompt engineering and retrieval refinement as effective mitigation strategies for improving response reliability. Farquhar et al. (2024) \cite{c8} introduced semantic entropy as a novel uncertainty estimation technique capable of detecting confabulations in LLM outputs without requiring domain-specific supervision. Their approach demonstrated strong generalization across unseen tasks and significantly improved the identification of unreliable responses. In addition, Ji et al. (2023) \cite{c9} provided a comprehensive survey of hallucination challenges across multiple Natural Language Generation (NLG) tasks, categorizing hallucinations into intrinsic and extrinsic forms while discussing mitigation techniques including reinforcement learning, data refinement, and architectural optimization. Collectively, these studies establish that effective hallucination mitigation requires robust retrieval grounding, uncertainty-aware reasoning, and context-sensitive evaluation mechanisms to ensure trustworthy response generation.

Pendyala and Hall (2024) \cite{c10} evaluated multiple LLMs, including Mistral, Llama, Orca, and Falcon, for misinformation detection across the COVID-19 and LIAR datasets, revealing that model effectiveness is highly dependent on dataset complexity and contextual understanding. Their use of explainable AI techniques, including LIME and SHAP, further highlighted the importance of interpretability in misinformation-sensitive applications. Adel and Alani (2025) \cite{c11} assessed the role of generative AI in systematic literature reviews and found that although LLMs improve efficiency in structured research tasks, hallucination rates remain substantially high in interpretative synthesis activities, thereby necessitating human oversight and trustworthy retrieval mechanisms. Similarly, Hu et al. (2025) \cite{c12} investigated the influence of LLM-generated fake news in recommendation ecosystems and demonstrated how synthetic misinformation accelerates truth decay by reducing the visibility and ranking of factual information. Han et al. (2024) \cite{c14} further revealed the vulnerability of medical LLMs to targeted misinformation attacks, showing that minimal modifications to model parameters can inject false biomedical knowledge while preserving overall model performance. These studies collectively highlight the growing risks associated with misinformation amplification, lack of explainability, and adversarial manipulation in modern LLM systems.

Although previous studies demonstrate the effectiveness of several AI models in diverse domains such as healthcare, manufacturing, responsible adoption in SME-specific environments remain comparatively underexplored. Table \ref{lit-summary} summarizes and compares the most relevant works in the literature. In addition, prior studies generally investigate isolated retrieval or mitigation techniques without systematically comparing vector and graph RAG mechanisms under the same experimental setting. SME-related datasets often contain loosely connected policy-oriented and cybersecurity-focused documents with limited relational structures, making the effectiveness of different retrieval strategies highly dependent on the nature of the underlying data. Furthermore, limited research has evaluated how different RAG approaches influence contextual relevance, hallucination risk, retrieval efficiency, and trustworthiness across multiple LLM architectures within SME environments. Therefore, this study conducts an extensive experimental analysis of vector and graph based RAG approaches utilizing SME-specific datasets across multiple state-of-the-art LLMs, including Mistral-7B, LLaMA, and Qwen. By evaluating retrieval effectiveness, contextual grounding, misinformation mitigation, and trustworthiness, this work aims to identify the most suitable retrieval strategy for reliable and context-aware AI systems in SMEs.

\begin{table*}[ht!]
\tiny
\centering
\caption{Key aspects in existing studies}
\label{lit-summary}
\renewcommand{\arraystretch}{1.3}
\begin{tabular}{|p{7cm}|p{2cm}|p{2cm}|p{2cm}|p{2cm}|}
\hline
\textbf{Features} & \textbf{Han et al. \cite{c15}} & \textbf{Pendyala et al. \cite{c12}} & \textbf{Mohammed et al. \cite{c16}} & \textbf{Zhang et al. \cite{c9}} \\
\hline
Use of Hybrid RAG (Knowledge Graph + Vector Retrieval) & \ding{55} & \ding{55} & \ding{55} & \checkmark \\
\hline
Real-Time Application in Business or Industry & \ding{55} & \ding{55} & \ding{55} & \ding{55} \\
\hline
Contextual Relevance for Specific Tasks (e.g., SMEs) & \ding{55} & \ding{55} & \ding{55} & \ding{55} \\
\hline
Use of Re-ranker or Knowledge Retrieval Enhancement & \ding{55} & \ding{55} & \checkmark & \ding{55} \\
\hline
Cross-LLM Evaluation for Misinformation Detection & \ding{55} & \checkmark & \ding{55} & \checkmark \\
\hline
Evaluation of Accuracy, Completeness, and Relevance & \checkmark & \checkmark & \checkmark & \checkmark \\
\hline
Robustness Against Data Variations (e.g., in SMEs) & \ding{55} & \ding{55} & \ding{55} & \ding{55} \\
\hline
Comprehensive Approach to Misinformation in Multiple Domains & \ding{55} & \ding{55} & \ding{55} & \ding{55} \\
\hline
Real-World Applicability in Business and Research Settings & \ding{55} & \ding{55} & \ding{55} & \ding{55} \\
\hline
Accuracy and Relevance in Task-Specific LLM Outputs & \ding{55} & \ding{55} & \ding{55} & \checkmark \\
\hline
\end{tabular}
\end{table*}

\section{Methodology} 
This study is primarily designed to investigate VectorRAG and GraphRAG approaches for SME-oriented applications. For this purpose, we collect a domain-specific corpus of SME-related PDF documents covering cybersecurity best practices, compliance requirements, and operational business knowledge. Both approaches are implemented and evaluated independently to provide a comprehensive understanding of their respective strengths, limitations, and applicability in SME contexts.

\begin{figure*}[htbp]
    \centering
    \includegraphics[width=\textwidth,height=0.9\textheight,keepaspectratio]{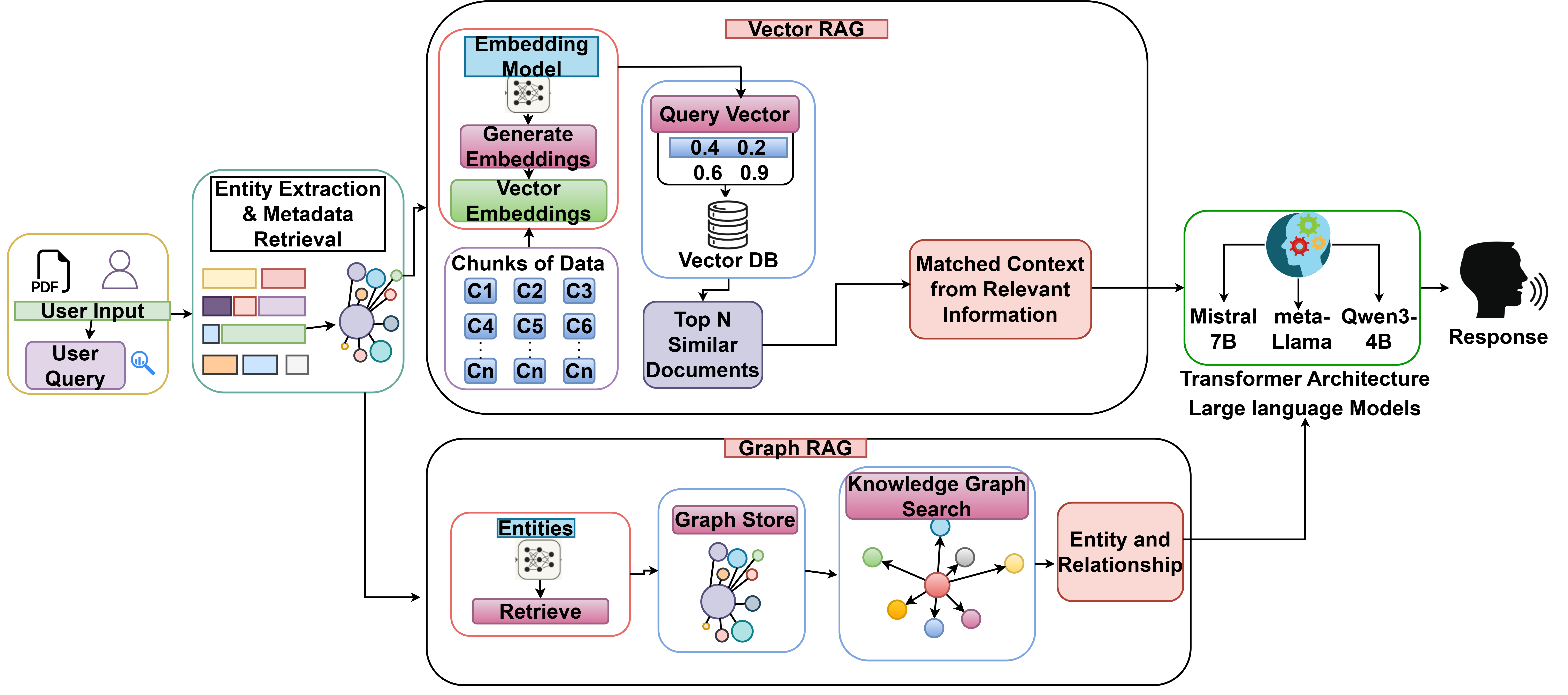}
    \caption{An illustration of the RAG-based system architecture}
    \label{fig:fig1}
\end{figure*}

For the vector RAG pipeline, the documents are preprocessed and segmented into smaller text chunks. These chunks are converted into dense vector embeddings using a pre-trained embedding model and stored in a vector database to support efficient semantic similarity retrieval. For the graph RAG pipeline, entities and relationships are extracted from the same corpus and stored in a knowledge graph constructed using Neo4j \cite{bratanic2024graph}. This structured representation enables retrieval based on entity relationships and supports reasoning over interconnected data. The overall workflow of the system, from input query to final response generation, is illustrated in Figure~\ref{fig:fig1}. The key modules of the system are presented and discussed in the following subsections.

\subsection{Dataset Acquisition}
The dataset used in this study is constructed from a collection of publicly available SME-related documents gathered from authoritative Australian government agencies, cybersecurity organizations, international policy institutions, industry reports, and peer-reviewed academic sources. The primary objective of the data acquisition process is to develop a domain-specific corpus capable of supporting Retrieval-Augmented Generation (RAG) for cybersecurity-focused SME knowledge retrieval and intelligent question answering. The documents collected focused on cybersecurity awareness, cyber resilience, governance frameworks, scam prevention, cloud security practices, financial resilience, operational challenges, business continuity, regulatory compliance, and the application of AI and LLMs within SME environments.

The final corpus consisted of a diverse collection of publicly available sources containing cybersecurity frameworks, national strategy documents, SME operational reports, industry surveys, academic research articles, and policy publications. Government and institutional publications are obtained from official Australian sources including the Australian Cyber Security Centre (ACSC), Australian Signals Directorate (ASD), Department of Home Affairs, Australian Competition and Consumer Commission (ACCC), Australian Information Security Association (AISA), and Australian Institute of Company Directors (AICD). These publications provided extensive information regarding cybersecurity governance, incident response, cyber hygiene, scam awareness, resilience strategies, risk mitigation, and SME-focused cybersecurity guidance.

The dataset also incorporated international policy reports, SME industry surveys, cybersecurity resilience studies, governance reports, and peer-reviewed academic literature to strengthen the broader contextual and operational coverage of the corpus. These documents contributed insights into SME digital transformation, financial resilience, cybersecurity preparedness, technology adoption, cyber insurance, governance challenges, cybersecurity legislation, AI-driven cybersecurity systems, Retrieval-Augmented Generation (RAG), and Large Language Model (LLM) applications relevant to SME environments. The integration of both policy-oriented publications and research-driven studies improved the diversity, contextual richness, and analytical depth of the dataset.

The documents are collected from publicly accessible repositories and official online platforms such as the Australian Cyber Security Centre (ACSC) (\url{https://www.cyber.gov.au}), Department of Home Affairs (\url{https://www.homeaffairs.gov.au}), Australian Signals Directorate (ASD) (\url{https://www.asd.gov.au}), Australian Competition and Consumer Commission (ACCC) (\url{https://www.accc.gov.au}), CPA Australia (\url{https://www.cpaaustralia.com.au}), Edith Cowan University Research Online (\url{https://ro.ecu.edu.au}), ResearchGate (\url{https://www.researchgate.net}), Elsevier (\url{https://www.sciencedirect.com}), and other institutional and academic repositories. All collected files are stored in Portable Document Format (PDF) to preserve structural consistency, metadata integrity, and textual fidelity.

\subsection{Vector RAG}
The Vector RAG process consists of several key steps, discussed below.

\begin{itemize}
    \item Chunking Process: The document is partitioned into overlapping chunks using a sliding window approach. The window size \( w \) is set to 700 tokens. The fragmentation process is mathematically described as follows:

    \[
    c_j = \text{split}(d_i, w, o) \tag{5}
    \]
    where \( d_i \) is the \( i \)-th document, \( w \) is the window size, and \( o \) is the overlap.

    Each chunk is stored with metadata, including its position in the document and relevant categories, for easy retrieval during later stages. This chunking approach ensures that the model can efficiently retrieve and process relevant information for answering queries.

    \item Vector Embeddings: The SME-related documents are converted into dense vector embeddings using a pre-trained transformer model. These embeddings represent the semantic content of the documents in a high-dimensional vector space, capturing the relationships and underlying meanings within the text.

    \item Vector Database: The embeddings are stored in a high-performance vector database, which is optimized for similarity search. The database allows for fast and efficient retrieval of document chunks based on their semantic similarity to a user query. This is critical for ensuring that the system retrieves the most relevant information. The retrieval process works by comparing the query's embedding with those stored in the database to find the most similar document chunks, which are then used for response generation.
    
    \item Top N similar Documents: In Vector RAG models, the top-N similarity documents refer to the most relevant documents retrieved based on their similarity to a given input query. This retrieval process involves transforming both the query and documents into vectors within a high-dimensional space using advanced embedding techniques, transformer-based models. The top-N documents are selected on the basis of their cosine similarity scores. These selected documents provide the necessary context for generating responses or augmenting the model's output, improving both the accuracy and relevance of the generated content. This mechanism enhances the performance of the RAG model by ensuring that the generated text is deeply informed by the most pertinent and contextually rich sources.
\end{itemize}

\subsubsection{Vector RAG Implementation}
This algorithm uses document embeddings and vector similarity search to retrieve relevant text chunks, which are supplied as context to an LLM for generating grounded, evidence-based answers. Algorithm \ref{vector_RAG_Algo} presents the detailed procedure for implementing the vector RAG approach.

\begin{algorithm}
\caption{Vector RAG}
\label{vector_RAG_Algo}
\begin{algorithmic}
\State \textbf{Input:} User query \( q \), document corpus \( \mathcal{D} = \{d_i\} \)
\State \textbf{Output:} Grounded answer \( \hat{y} \) generated by an instruction-tuned LLM using retrieved evidence.
\State \textbf{Hyperparameters:} Chunk size \( w = 700 \), overlap \( o = 100 \), max new tokens \( = 512 \), temperature \( = 0.7 \)

\State \textbf{Step 1: Corpus Ingestion and Preprocessing}
\For{each document \( d_i \) in \( \mathcal{D} \)}
    \State Load the document \( d_i \)
    \State Normalize text (remove control chars, collapse whitespace)
    \State Split into overlapping chunks \( c_j \) with window length \( w \) and overlap \( o \)
    \EndFor

\State \textbf{Step 2: Embedding and Indexing}
\For{each chunk \( c_j \)}
    \State Compute embedding \( \mathbf{e}_j = E(c_j) \in \mathbb{R}^m \)
    \State Upsert \( (\mathbf{e}_j, \text{metadata}_j) \) into Qdrant (or FAISS fallback)
\EndFor

\State \textbf{Step 3: Optional Domain Adaptation (PEFT SFT)}
\If{training data \( \mathcal{T} \) is provided}
    \State Format instruction pairs
    \State Fine-tune Mistral-7B/LLama/Qwen3 using PEFT/LoRA to obtain \( \theta' \)
\Else
    \State Use pre-trained Mistral-7B/LLama/Qwen3 (\( \theta \))
\EndIf

\State \textbf{Step 4: Retrieval at Inference Time}
\State Compute query embedding \( \mathbf{e}_q = E(q) \)
\State Retrieve top-k chunks \( \{c_{j_1},\ldots,c_{j_k}\} \) from Index

\State \textbf{Step 5: Prompt Construction}
\State Build structured prompt \( p = S(q, \{c_{j_r}\}) \)

\State \textbf{Step 6: Generation}
\State Generate the answer \( \hat{y} = f_{\theta^*}(p; \text{temperature}, \text{max\_new\_tokens}) \)
\State \Return \( \hat{y} \) with citations from metadata
\end{algorithmic}
\end{algorithm}

Vector RAG \cite{c16} enhances LLM performance by combining vector search with external unstructured text sources. The workflow starts when a user asks a question about information that lies outside the LLM’s original training data. The external documents are first split into smaller chunks. Each chunk is then transformed into a vector using an embedding model, and all vectors are stored in a vector database. When a new query arrives, it too is converted into an embedding. The retrieval system compares this query vector with those in the database, using similarity scoring to find and rank the chunks that best match the user’s intent. The highest-scoring chunks are collected and passed to the LLM along with the original query. Equipped with this additional context, the LLM can produce a response that reflects the most relevant and up-to-date information available. This vector method significantly improves answer accuracy and relevance. However, because it relies exclusively on unstructured text retrieval, it may fall short when handling queries that demand structured reasoning or a deeper understanding of domain-specific relationships. Table \ref{vec-para} summarizes the parameters used in the vector RAG setup.

\begin{table}[ht]
\centering
\caption{Vector RAG setup parameters.}
\label{vec-para}
\begin{tabular}{ll}
\hline
\textbf{Configuration} & \textbf{Details} \\
\hline
Language Model (LLM) & Mistral7B, Llama, Qwen3 \\
Model for Embedding & text-embedding-3-small \\
Pipeline for RAG & Langchain \\
Size of Chunks & 700 \\
Overlap between Chunks & 100 \\
Maximum Token Count for Output & 512 \\
Temperature & 0.7 \\
Vector Store & Qdrant (FAISS fallback) \\
\hline
\end{tabular}
\end{table}

\subsection{Graph RAG}
In the graph RAG process, the first step involves entity extraction, where key entities from the input query are identified. Next, document retrieval takes place, where relevant documents are retrieved through a vector search, matching the query to the closest document embeddings. These entities and relationships are then stored in a knowledge graph, creating a structured repository of information. The knowledge graph is subsequently searched to find related entities and relationships, providing additional context. Finally, relevant entities and relationships are extracted from the graph, which are used to augment the generation process, ensuring that the output is contextually rich and more accurate.
\begin{itemize}
 
\item Document Retrieval: After extracting the entities, the system retrieves relevant documents using vector search. The query is converted into a vector embedding using a pre-trained model, and these embeddings are compared with stored document embeddings in the database. This comparison is based on semantic similarity, enabling the system to retrieve document chunks that are contextually relevant, even if there are no exact keyword matches.

\item Knowledge Graph Storage: The retrieved documents, along with the identified entities and their relationships, are stored in a knowledge graph. This graph organizes the entities and their interconnections in a structured manner, allowing for easy and efficient access to related information. The knowledge graph serves as a dynamic repository of relevant data that can be used to enhance the response generation process.

\item Graph Search and Knowledge Integration: Once the knowledge is stored in the graph, a search is conducted to find relevant entities and relationships that provide additional context to the query. These results are then integrated into the response generation process, enriching the model's output. This combination of the model's pre-trained knowledge and the external, graph information allows the system to generate responses that are more informed, specific, and contextually relevant.
\end{itemize}

\subsubsection{Graph RAG Implementation}
Graph RAG Algorithm extracts query intent, retrieves structured facts from a knowledge graph using Cypher queries, and generates answers through an LLM using graph-derived evidence. Algorithm \ref{Graph_RAG_Algo} presents the detailed procedure for implementing the GraphRAG approach.
\begin{algorithm}
\caption{Graph RAG}
\label{Graph_RAG_Algo}
\begin{algorithmic}
\State \textbf{Input:} User query \( q \)
\State \textbf{Output:} Answer \( \hat{y} \) generated by selected LLM using retrieved facts from the knowledge graph
\State \textbf{Hyperparameters:} Max facts \( \text{max\_facts} = 25 \), max new tokens \( = 128 \), max sentences \( = 4 \)
\State \textbf{LLMs Supported:} \{MistralAI, Qwen3, LLaMA\}

\State

\State \textbf{Step 1: Intent Extraction using Selected LLM}
\State Choose LLM backend:
\State \quad \( \text{LLM} \in \{\text{Mistral7B}, \text{Qwen3}, \text{LLaMA}\} \)
\State Query LLM to extract intent:
\State \quad - Identify candidate \textit{head\_type}, \textit{tail\_type}
\State \quad - Extract \textit{relations}, \textit{head\_keywords}, \textit{tail\_keywords}
\State \textit{Return: GraphIntent \( G_{\text{intent}} \)}

\State

\State \textbf{Step 2: Build Cypher Query from Intent}
\State Construct Cypher query using:
\State \quad - Entity types: \( \text{head\_type}, \text{tail\_type} \)
\State \quad - Relations: \( \text{relations} \)
\State \quad - Keywords: \( \text{head\_keywords}, \text{tail\_keywords} \)
\State \textit{Return: Cypher query \( C_{\text{query}} \)}

\State

\State \textbf{Step 3: Execute Cypher Query on Knowledge Graph}
\State Execute \( C_{\text{query}} \) against Neo4j:
\State \quad - Retrieve facts \( \{ \text{head}, \text{relation}, \text{tail} \} \)
\State \textit{Return: Facts \( F_{\text{facts}} \)}

\State

\State \textbf{Step 4: Relax Query and Retry (if needed)}
\If{ \( F_{\text{facts}} = \emptyset \) }
    \State Relax constraints:
    \State \quad - Remove type restrictions
    \State \quad - Allow fuzzy matching with keywords
    \State Re-run Cypher query
    \State \textit{Return: Relaxed Facts \( F_{\text{facts}} \)}
\EndIf

\State

\State \textbf{Step 5: Fallback Retrieval}
\If{ \( F_{\text{facts}} = \emptyset \) }
    \State Perform broad fallback graph traversal
    \State \textit{Return: Fallback Facts \( F_{\text{fallback}} \)}
\EndIf

\State

\State \textbf{Step 6: Answer Generation using Selected LLM}
\State Format retrieved facts as:
\State \quad \( \{ \text{head} -- \text{relation} --> \text{tail} \} \)
\State Build structured LLM prompt
\State Generate answer using:
\State \quad \( \text{LLM} \in \{\text{Mistral7B}, \text{Qwen3}, \text{LLaMA}\} \)
\State \textit{Return: Raw Answer \( \hat{y} \)}

\State

\State \textbf{Step 7: Post-process Answer}
\State Trim whitespace, normalize formatting
\State Keep maximum 4 sentences

\State

\Return Final Answer \( \hat{y} \)

\end{algorithmic}
\end{algorithm}

Graph RAG \cite{c17} enhances retrieval by using a domain-specific knowledge graph (KG) to supply structured information. Like vector RAG, it starts with a user query, but instead of searching through text embeddings, the system navigates the KG to find relevant entities and their relationships. The retrieval module pulls out a subgraph—nodes and edges that are semantically connected to the query.

This subgraph is then converted into a format the LLM can understand, such as triples or graph embeddings. With this structured context, the LLM can generate answers that reflect precise domain relationships, improving accuracy for queries that depend on specialized knowledge. The main limitation is the KG itself: if it is incomplete or too coarse, the model’s answers may miss important details. Table \ref{graph-para} summarizes the parameters used in the graph RAG setup.

\begin{table}[ht]
\centering
\caption{GraphRAG configuration parameters.}
\label{graph-para}
\begin{tabular}{ll}
\hline
\textbf{Configuration} & \textbf{Details} \\
\hline
Language Model (LLM) & Mistral7B, LLaMA, Qwen3 \\
RAG Framework & LangChain \\
Knowledge Graph Database & Neo4j \\
Maximum Retrieved Facts & 25 \\
Maximum Output Tokens & 128 \\
Maximum Answer Length & 4 sentences \\
Total Number of Triples & 2563 \\
Total Number of Nodes & 1896 \\
Total Number of Edges & 2135 \\
\hline
\end{tabular}
\end{table}

\subsection{LLM Model Selection}

This work employs multiple state-of-the-art LLMs to answer queries related to SMEs' cybersecurity practices, business operations, and compliance. The models selected for this work include:

\begin{itemize}
    \item Mistral-7B: A 7-billion parameter transformer model \cite{c18}, fine-tuned to enhance its understanding of domain-specific queries related to SMEs, particularly focusing on cybersecurity and business management. The model was optimized with 4-bit quantization, a technique that reduces memory usage while maintaining performance, enabling efficient processing on available hardware. The Mistral-7B architecture is based on a dense transformer network and leverages the self-attention mechanism for encoding input data \cite{c19}. The model is fine-tuned using domain-specific datasets, which helps improve the accuracy and relevance of the model's responses in SME-related domains. One of the key features of Mistral-7B is its ability to handle long-range dependencies in text, making it ideal for tasks requiring understanding of complex regulatory frameworks or cybersecurity protocols. The transformer architecture used in Mistral-7B is defined by the following equation for self-attention:

    \[
    \text{Attention}(Q, K, V) = \text{softmax}\left(\frac{QK^T}{\sqrt{d_k}}\right)V \tag{1}
    \]
    where:
    - \( Q \) is the query matrix,
    - \( K \) is the key matrix,
    - \( V \) is the value matrix,
    - \( d_k \) is the dimension of the key vectors.

    This equation computes the attention score between each pair of input tokens and then uses these scores to weight the value vectors, allowing the model to focus on relevant parts of the input. Additionally, the model's parameters are quantized to 4 bits, which allows it to operate efficiently on hardware with limited memory resources while maintaining the integrity of the model's outputs. Quantization is the process of reducing the precision of the model's weights from 32 bits (floating point) to 4 bits, effectively compressing the model size. The equation for quantization in this context is given by:

    \[
    w_q = \text{round}\left(\frac{w}{\Delta}\right) \tag{2}
    \]
    where:
    - \( w \) is the original weight,
    - \( \Delta \) is the scaling factor for quantization,
    - \( w_q \) is the quantized weight.

    This reduces the model’s memory footprint without significantly degrading performance.

    \item Qwen3-4B: A model designed to handle a variety of business-related queries \cite{c20}, including customer feedback, business professionalism, and data protection. Qwen3-4B is well-suited for generating structured, coherent responses based on complex input queries. The Qwen3-4B architecture leverages a multi-layer perceptron (MLP) approach in conjunction with a transformer-based architecture to handle business-specific tasks \cite{c21}. The following equation defines the forward pass for Qwen3-4B:

    \[
    \hat{y} = \text{MLP}\left(\text{Transformer}(X)\right) \tag{3}
    \]
    where:
    - \( X \) represents the input data (e.g., customer feedback, queries),
    - \(\text{Transformer}(X)\) is the transformer encoding process,
    - \( \hat{y} \) is the output prediction (e.g., response to a query).
    
    The MLP helps transform the output from the transformer layers into a more structured response suitable for business-related tasks.

    \item Llama-3.1: The Llama-3.1 model \cite{c22} is a powerful language model known for its efficiency in generating responses that are both clear and concise, making it particularly useful for technical topics like cybersecurity and business compliance. Llama-3.1 has been trained to handle more specific, detailed queries, and its flexibility in natural language generation makes it an ideal choice for this study's objectives \cite{c23}. The architecture of Llama-3.1 follows a similar transformer-based framework but incorporates a feed-forward network (FFN) that operates after the attention mechanism to process information more efficiently. The equation for the feed-forward network is given by:

    \[
    \text{FFN}(x) = \text{ReLU}(xW_1 + b_1)W_2 + b_2 \tag{4}
    \]
    where:
    - \( x \) is the input to the FFN,
    - \( W_1, W_2 \) are weight matrices,
    - \( b_1, b_2 \) are bias terms,
    - \( \text{ReLU} \) is the activation function applied element-wise to the input.

    This FFN operation enables Llama-3.1 to effectively model more complex relationships in the data, such as technical terms in cybersecurity and business compliance.
\end{itemize}

To achieve our goal, these popular LLM models are implemented and evaluated on the basis of their ability to process SME-related queries and provide relevant responses derived from the gathered documents.

\section{Experimental Evaluation}
This section presents an extensive experimental study conducted across various evaluation dimensions. The following subsections provide a detailed analysis and discussion of the experimental results.
\subsection{Evaluation Metrics}
In this study several metrics are used to assess response quality, factual grounding, retrieval effectiveness, and computational efficiency. The outcome is calculated based on the generated answer, retrieved context, and reference answer from the SME evaluation dataset.

\begin{itemize}
    \item \textbf{METEOR:} METEOR \cite{c15} is calculated by comparing the generated answer with the reference answer using exact word matching, stemming, and synonym matching. Higher METEOR scores indicate greater semantic similarity between the generated and expected responses. 

    \item \textbf{Faithfulness:} Faithfulness is calculated by measuring whether the information contained in the generated answer is directly supported by the retrieved context. Responses that closely reflected the retrieved evidence achieved higher faithfulness scores. 

    \item \textbf{Hallucination Risk:}  Hallucination Risk \cite{c15} is calculated as the proportion of generated content that could not be verified from the retrieved documents. Lower values indicate a lower likelihood of producing unsupported or misleading information.

    \item \textbf{Context Relevance:} Context Relevance is calculated by evaluating the semantic similarity between the user query and the retrieved document chunks. Higher scores indicate that the retrieval mechanism successfully identified contextually relevant information. 

    \item \textbf{Completeness:} Completeness \cite{c15} \cite{c16} is calculated by measuring the extent to which the generated response covered all important information contained in the reference answer. Responses containing all key facts achieved higher completeness scores.

    \item \textbf{ROUGE-L Recall:} ROUGE-L Recall is calculated using the longest common subsequence (LCS) between the generated answer and the reference answer. Higher scores indicate better content coverage and alignment with the reference response. 

    \item \textbf{Cosine Similarity:} Cosine Similarity \cite{c12} is calculated by measuring the cosine of the angle between the embedding vectors of the generated answer and the reference answer. It evaluates semantic similarity regardless of exact word overlap, and higher values indicate greater semantic alignment. 

    \item \textbf{F1-BERT:} F1-BERT is calculated using contextual embeddings generated by a pre-trained BERT model to measure semantic similarity between the generated and reference answers. The metric combines precision and recall at the token embedding level to produce an F1 score.

\end{itemize}

\subsection{Effectiveness Analysis}

\begin{figure*}[htbp]
    \centering
    \includegraphics[width=\textwidth]{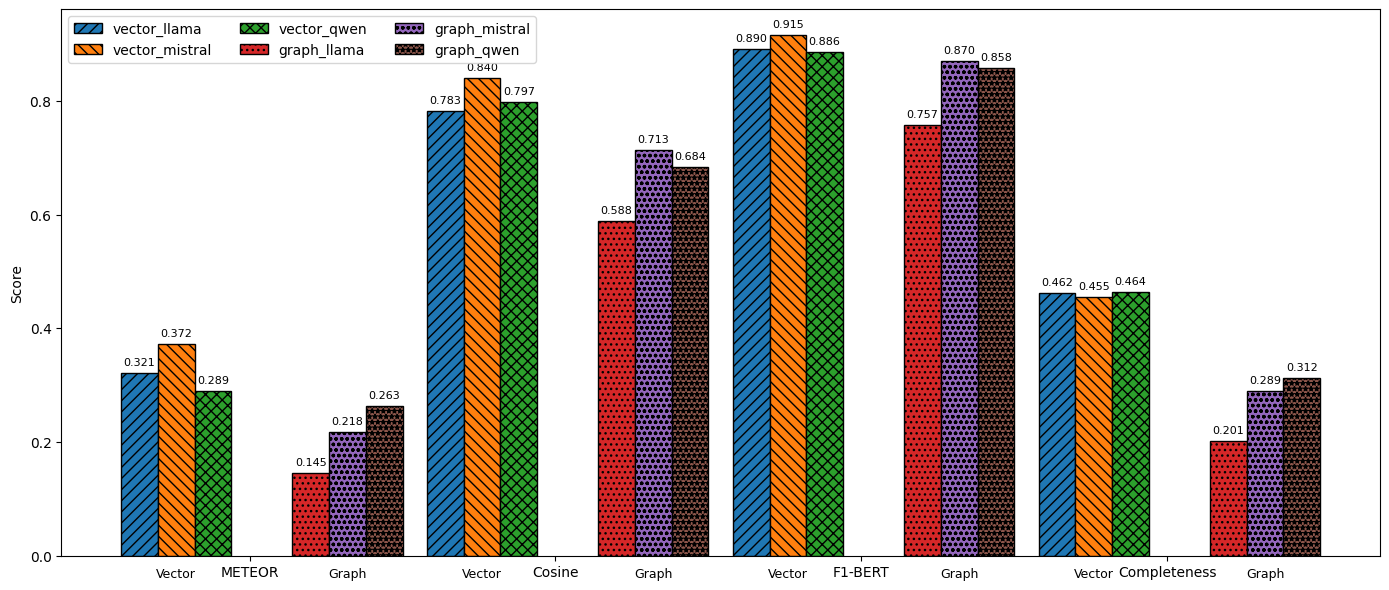}
    \caption{Effectiveness of LLMs: Vector RAG vs. Graph RAG}
    \label{fig:fig4}
\end{figure*}

Figure~\ref{fig:fig4} presents a comparative evaluation of VectorRAG and GraphRAG pipelines across three LLMs (Llama, Mistral, and Qwen) using METEOR, Cosine Similarity, F1-BERT, and Completeness metrics. Overall, the VectorRAG pipelines consistently outperform their GraphRAG counterparts across all evaluation measures.

For METEOR if the reference answer is "update software regularly" and the generated answer is "perform regular software updates" it matches the answers and METEOR gives a high score. Vector RAG models achieved substantially higher scores than GraphRAG. The best performance was obtained by Vector-Mistral (0.372), followed by Vector-Llama (0.321) and Vector-Qwen (0.289), whereas GraphRAG scores ranged from 0.145 to 0.263. This indicates that vector-based retrieval generated responses with greater lexical similarity to the reference answers.

For cosine similarity, if the reference answer is ``update software regularly'' and the generated answer is ``perform regular software updates,'' the embeddings capture similar meanings, resulting in a high cosine similarity score. Conversely, semantically unrelated responses yield lower scores. A similar trend is observed for Cosine Similarity, where VectorRAG achieved scores between 0.783 and 0.840, significantly exceeding the GraphRAG range of 0.588–0.713. The highest score was recorded by Vector-Mistral (0.840), suggesting that vector retrieval provided more semantically aligned contextual information to the language models.

For faithfulness if the generated answer states that small businesses enable multi-factor authentication, perform regular software updates, and back up important business data, a generated answer that includes only these recommendations would receive a high faithfulness score because all information is directly supported by the retrieved evidence.

For F1-BERT, which measures contextual and semantic similarity for example, if the reference answer is ``enable multi-factor authentication (MFA)'' and the generated answer is ``use MFA for account security,'' the wording differs but the semantic meaning remains similar, resulting in a high F1-BERT score. In contrast, responses containing unrelated or incomplete information receive lower scores. In this scenario, Vector RAG again demonstrated superior performance. Vector-Mistral achieved the highest score (0.915), followed by Vector-Llama (0.890) and Vector-Qwen (0.886). In contrast, GraphRAG models obtained lower scores between 0.757 and 0.870. Although Graph-Mistral performed competitively (0.870), it remained below all VectorRAG configurations.

Regarding Completeness if the ground truth includes three key security measures (multi-factor authentication, software updates, and data backups), but the generated answer includes only two of them (MFA and software updates), the completeness score is computed as $2/3 = 0.67$. If all three key points are included, the completeness score becomes $3/3 = 1.0$. VectorRAG maintained a clear advantage, with scores around 0.455–0.464 compared to 0.201–0.312 for GraphRAG. This suggests that vector-based retrieval retrieved more comprehensive information, enabling the models to produce more complete answers.

Table \ref{tab: Pipeline Evaluation Metrics} summarizes the performance of the six RAG pipelines using additional retrieval and generation quality metrics, including Context Relevance, Completeness, METEOR, Faithfulness, Hallucination Risk, and ROUGE-L Recall.
The VectorRAG pipelines achieved the highest Context Relevance scores, with all three models obtaining 0.9435, demonstrating their ability to retrieve highly relevant information from the document corpus. In contrast, GraphRAG pipelines achieved substantially lower scores, ranging from 0.2314 to 0.2431, indicating weaker retrieval relevance.
For Completeness, Vector-Qwen achieved the highest score (0.4639), closely followed by Vector-Llama (0.4624) and Vector-Mistral (0.4547). GraphRAG pipelines produced notably lower completeness values, with Graph-Qwen (0.3119) outperforming the other graph-based configurations.

The Hallucination Risk metric reveals a significant advantage for VectorRAG. The vector-based pipelines maintained extremely low hallucination scores (0.0028–0.0073), whereas GraphRAG exhibited considerably higher risks (0.0881–0.1053). Similar patterns were observed for ROUGE-L Recall, reflecting stronger alignment between retrieved content and generated responses in the VectorRAG configurations.

\begin{table*}[htbp]
\caption{LLM Pipeline Evaluation Metrics}
\label{tab: Pipeline Evaluation Metrics}
\centering
\tiny
\begin{tabular}{lcccccc}
\hline
\textbf{Pipeline} & \textbf{Context Relevance} & \textbf{Completeness} & \textbf{Meteor} & \textbf{Faithfulness} & \textbf{Hallucination Risk} & \textbf{RougeL Recall} \\
\hline
vector\_llama   & 0.9435 & 0.4624 & 0.321 & 0.2839 & 0.0028 & 0.0028 \\
vector\_mistral & 0.9435 & 0.4547 & 0.372 & 0.1968 & 0.0033 & 0.0033 \\
vector\_qwen    & 0.9435 & 0.4639 & 0.289 & 0.2733 & 0.0073 & 0.0073 \\
graph\_llama    & 0.2351 & 0.2013 & 0.145 & 0.3029 & 0.0881 & 0.0881 \\
graph\_mistral  & 0.2314 & 0.2892 & 0.218 & 0.3499 & 0.1053 & 0.1053 \\
graph\_qwen     & 0.2431 & 0.3119 & 0.263 & 0.1598 & 0.1034 & 0.1034 \\
\hline
\end{tabular}
\label{tab:metrics}
\end{table*}

Collectively, these findings indicate that VectorRAG provides stronger retrieval quality and response generation performance than Graph RAG for the evaluated SME cybersecurity question-answering tasks. Although GraphRAG leverages structured knowledge representation and relationship-based reasoning, its lower retrieval coverage and semantic alignment resulted in reduced answer quality across most metrics. This outcome may be partially attributed to the relatively limited dataset size, consisting of only 31 PDF documents. Since GraphRAG relies on rich inter-document entity relationships and graph connectivity to maximize retrieval effectiveness, its advantages may not be fully realized in smaller corpora. Consequently, the dense semantic retrieval capabilities of VectorRAG proved more effective for the current experimental setting, yielding higher relevance, completeness, and overall response quality.

\subsection{Response Generation using Vector RAG}
\begin{figure*}[htbp]
    \centering
    \includegraphics[width=.8\textwidth]{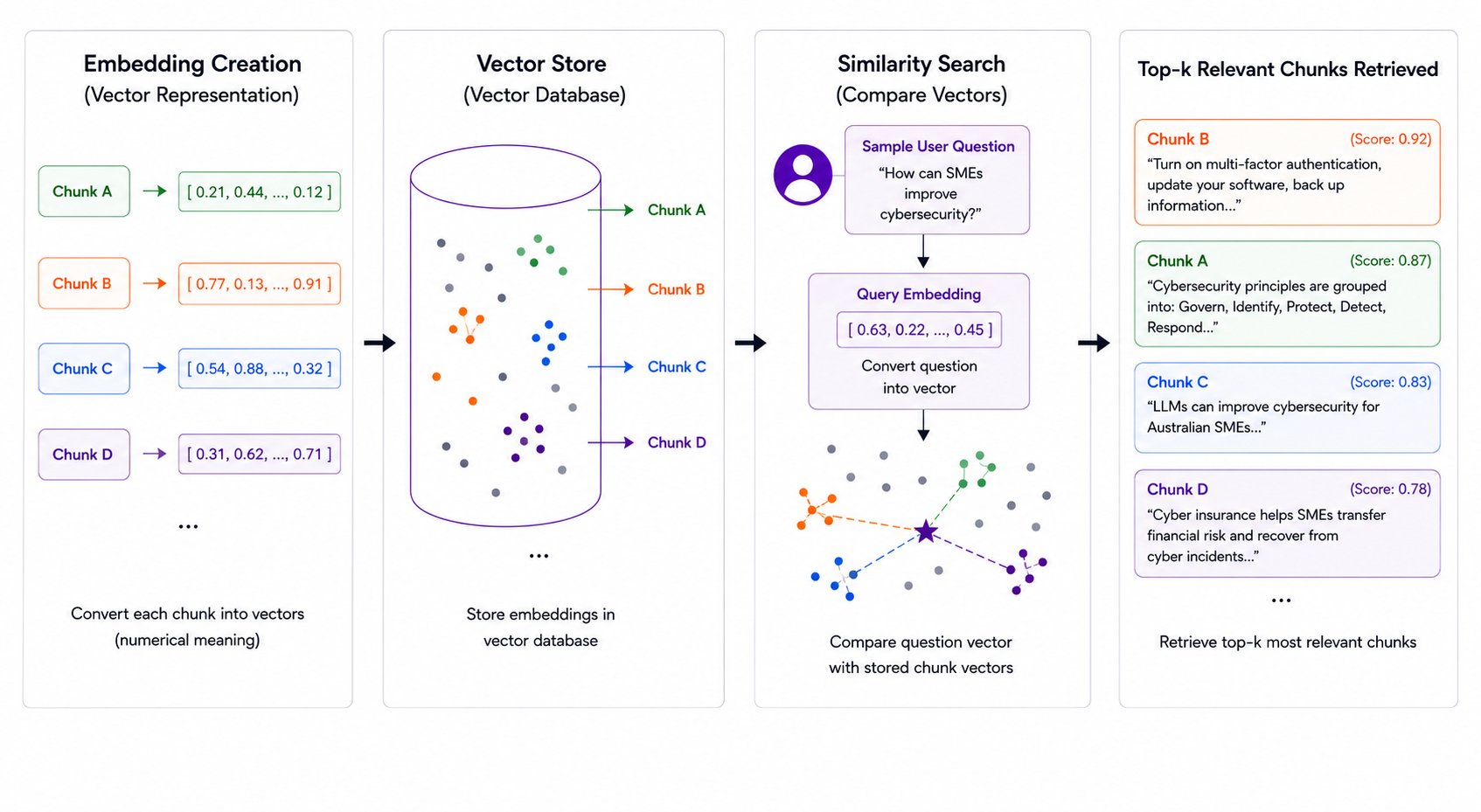}
    \caption{Response Generation using Vector RAG}
    \label{fig:fig8}
\end{figure*}

Figure~\ref{fig:fig8} presents an example of the RAG workflow used in the proposed framework. In this example, the sample user query is ``How can SMEs improve cybersecurity?''; and the generated Vector RAG response is "SMEs can improve cybersecurity by enabling multi-factor authentication (MFA), updating software regularly, backing up information, and following cybersecurity governance frameworks namely the ISM principles (Govern, Identify, Protect, Detect, Respond) to manage risks systematically." The query is converted into a vector embedding and compared with stored document embeddings in the vector database to retrieve the top-$k$ most relevant chunks, as shown in Figure~\ref{fig:fig8}. These retrieved contextual chunks are then supplied to the LLM to generate the final response. This process enables the model to produce more accurate, context-aware, and reliable cybersecurity guidance while reducing misinformation and hallucinated responses. 

\subsection{Constructed Knowledge Graph}
\begin{figure*}[htbp]
    \centering
    \includegraphics[width=0.8\textwidth]{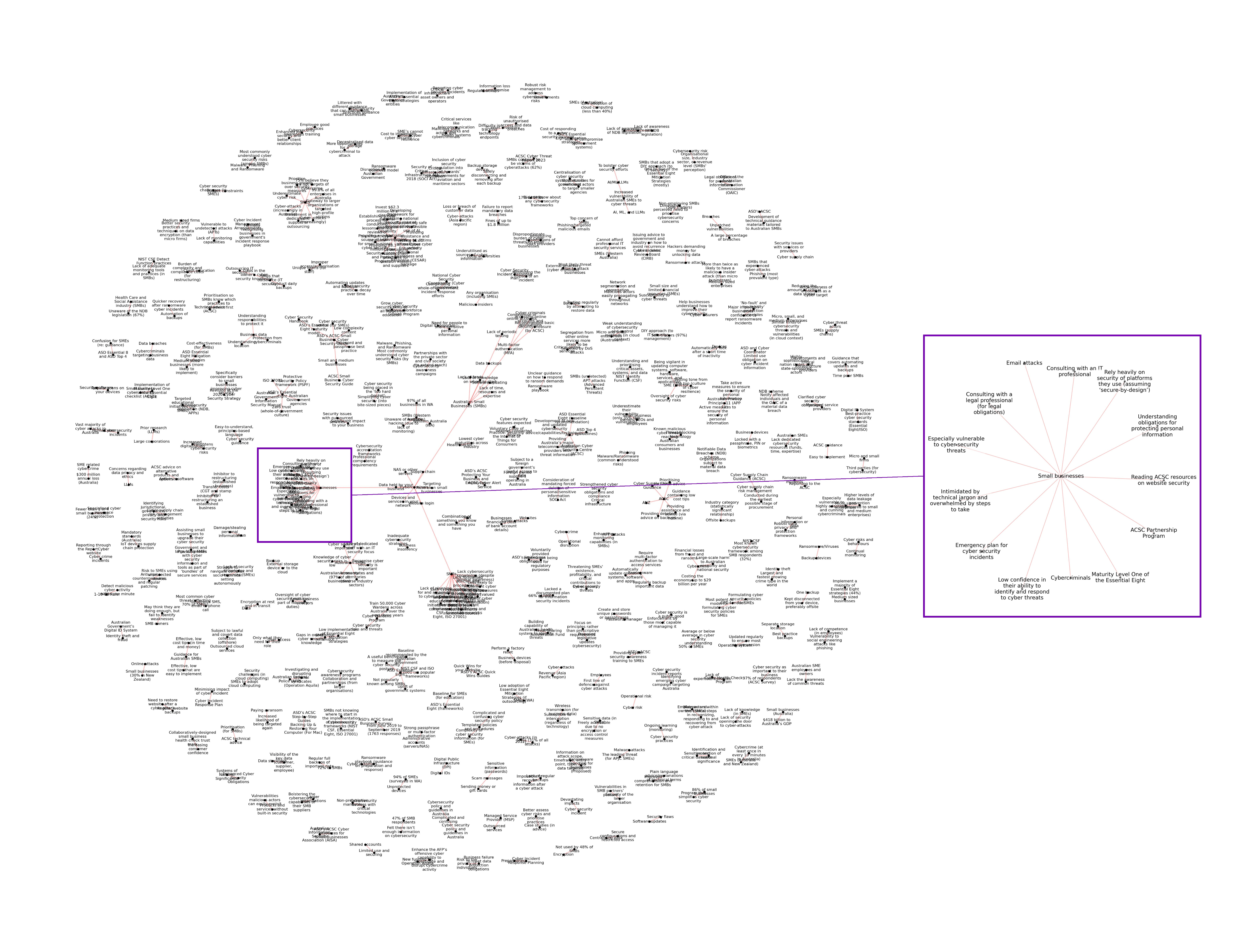}
    \caption{Illustration of constructed KG for SMEs.}
    \label{fig:fig6}
\end{figure*}

Figure~\ref{fig:fig6} illustrates a complex network of interconnected topics, with "SME" as the central node. This visualization represents the relationships between various concepts, including cybersecurity concerns, such as "email attacks," "vulnerable to threats," and "consulting with IT experts," all of which are essential for understanding the broader context of cybersecurity risks faced by small businesses. 

This structure exemplifies the core functionality of graph RAG, which relies on an extensive knowledge base of interconnected data points \cite{c24}. By leveraging these links, RAG models can retrieve and synthesize relevant, context-specific information to enhance the quality of their responses. In this case, the model would retrieve pertinent data from various cybersecurity nodes, ensuring that responses to queries about small business security are not only comprehensive but also informed by the most up-to-date and authoritative sources.

The interconnected nature of the graph underscores the importance of contextual relevance in RAG-based models. As the model retrieves information from these interconnected sources, it can generate more precise, actionable, and reliable answers. This process significantly enhances the trust and explainability of the model, as users can trace the connections between the retrieved information and the final response, thus ensuring that the provided guidance is well-supported by authoritative knowledge. The application of such an approach in the context of small business cybersecurity offers a clear example of how RAG models can improve decision-making by offering detailed and contextually enriched insights.

Moreover, the highlighted subgraph in Figure~\ref{fig:fig6} demonstrates how specific cybersecurity concepts form tightly connected semantic communities within the overall knowledge graph. For instance, nodes associated with "email attacks," "emergency plans for cyber security incidents," and "consulting with IT professionals" collectively indicate that SMEs primarily associate cybersecurity preparedness with reactive defense strategies and external expertise. Similarly, concepts such as "low confidence in identifying cyber threats" and "reliance on online security resources" reveal existing awareness and capability gaps among small businesses. These clustered relationships validate the effectiveness of the graph retrieval mechanism, as semantically related cybersecurity entities are naturally grouped together, enabling the RAG framework to retrieve richer contextual knowledge during inference. Consequently, the graph structure not only improves information retrieval accuracy but also enhances explainability by explicitly revealing the semantic pathways connecting cybersecurity risks, mitigation practices, and organizational awareness factors.

\begin{figure*}[htbp]
    \centering

    \begin{subfigure}{0.32\textwidth}
        \centering
        \includegraphics[width=\linewidth]{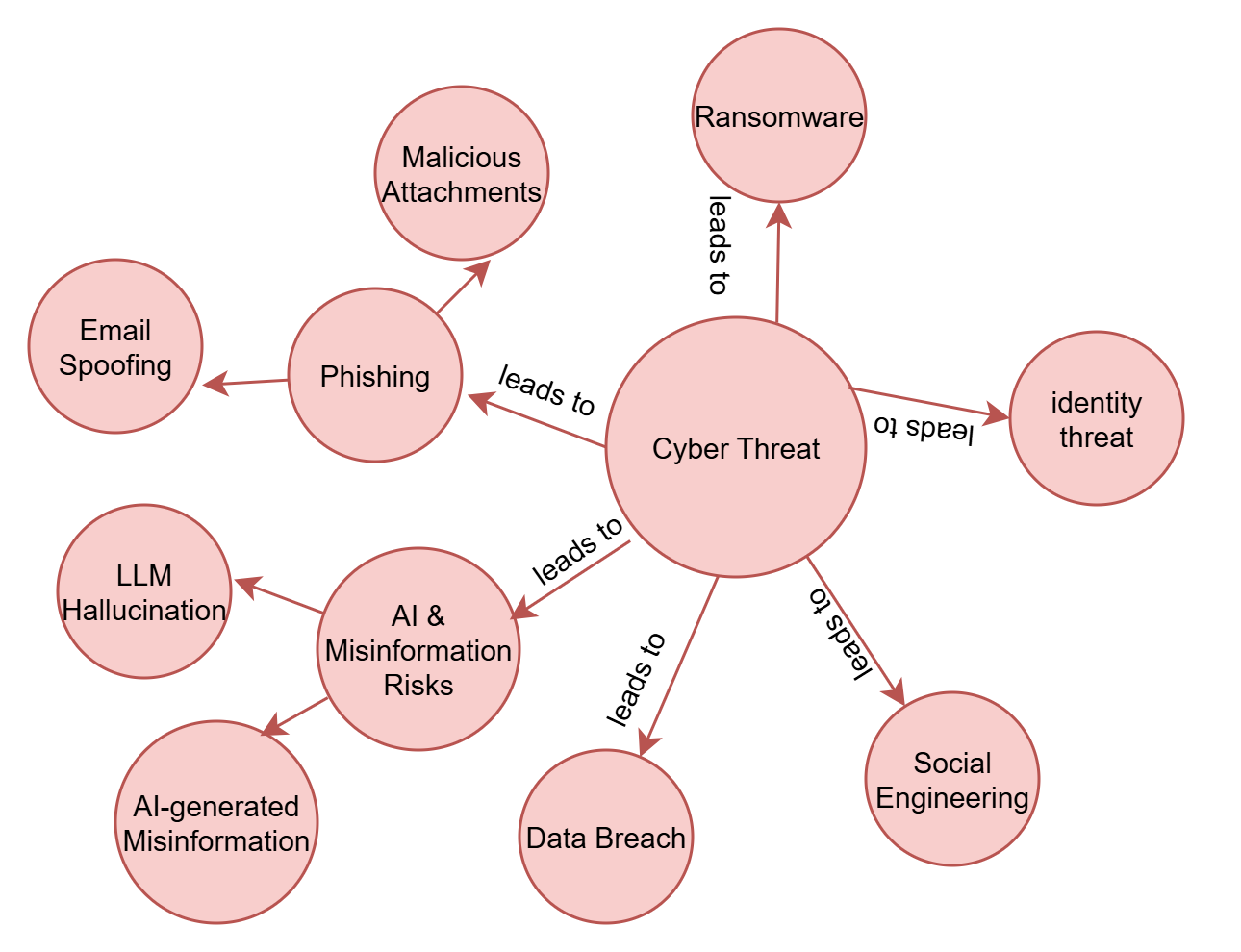}
        \caption{}
        \label{fig:kg_a}
    \end{subfigure}
    \hfill
    \begin{subfigure}{0.32\textwidth}
        \centering
        \includegraphics[width=\linewidth]{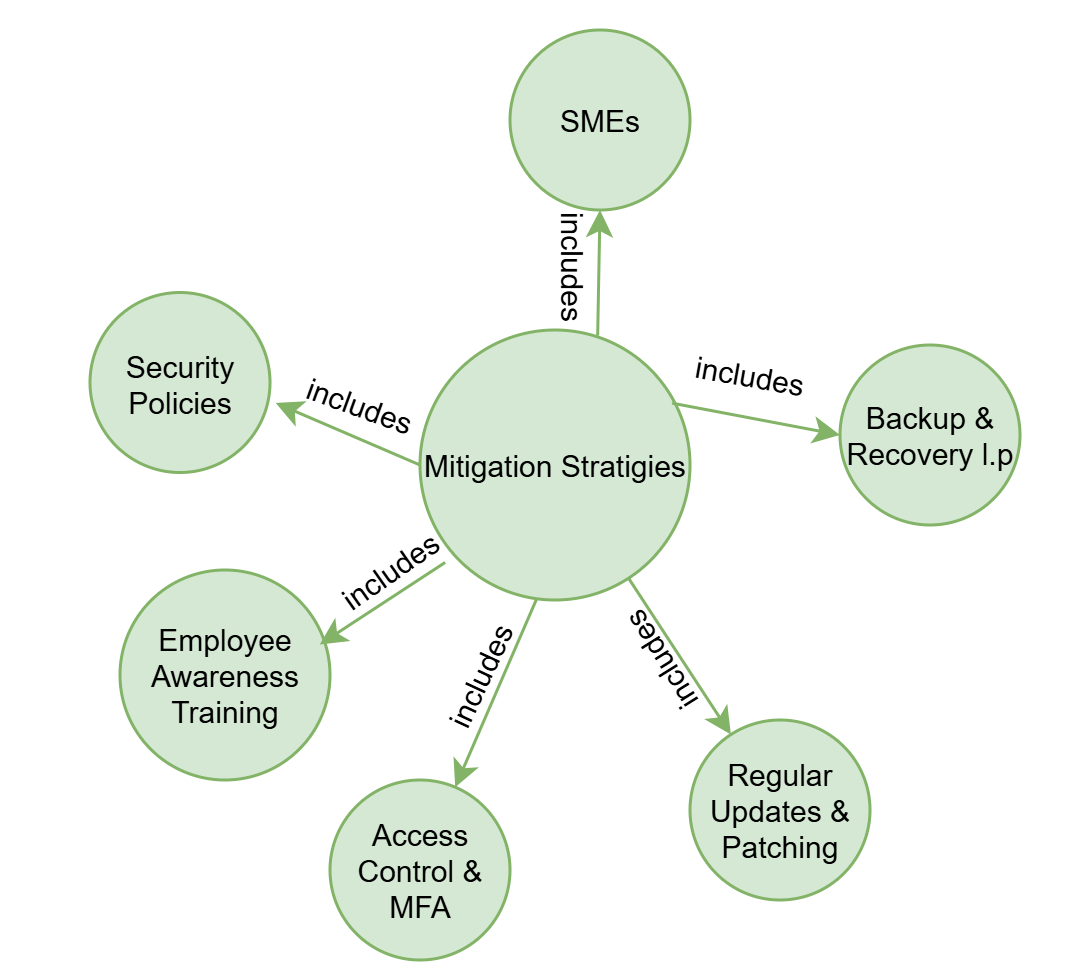}
        \caption{}
        \label{fig:kg_b}
    \end{subfigure}
    \hfill
    \begin{subfigure}{0.32\textwidth}
        \centering
        \includegraphics[width=\linewidth]{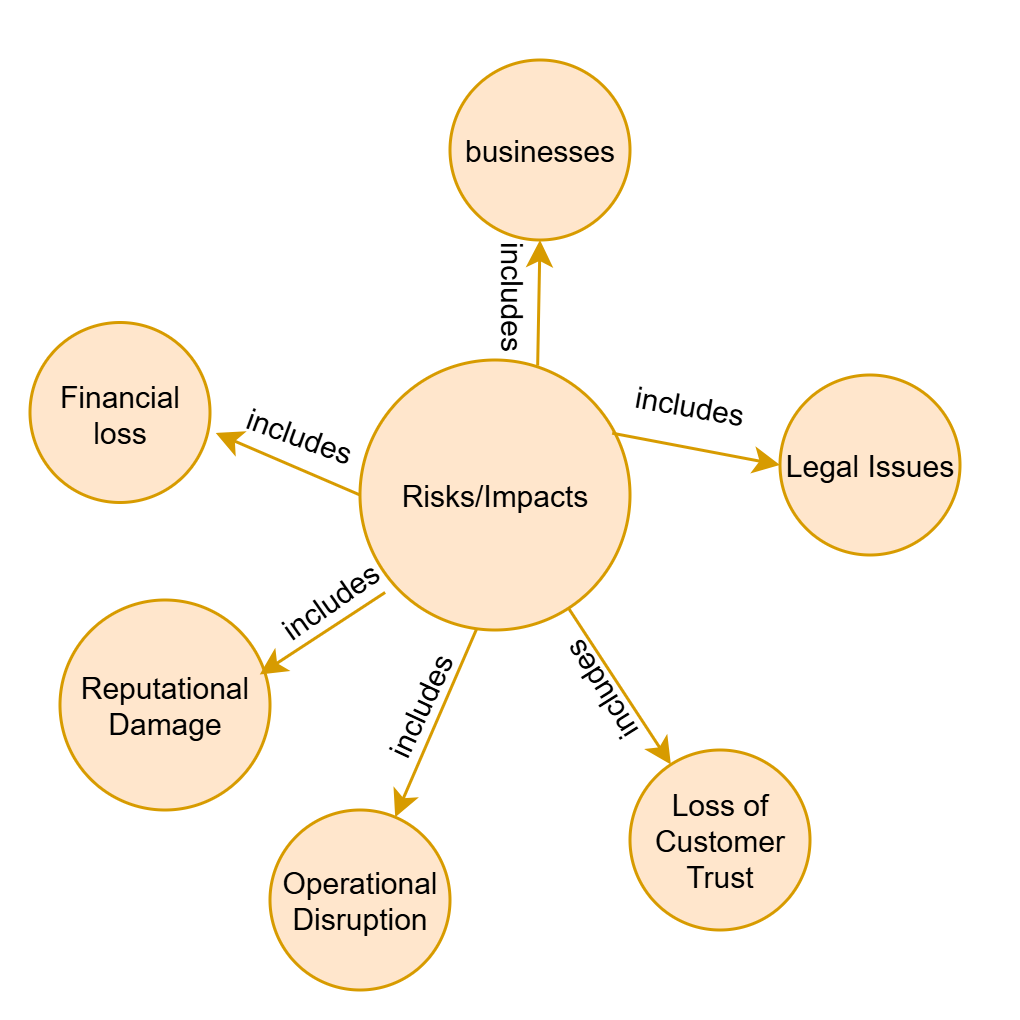}
        \caption{}
        \label{fig:kg_c}
    \end{subfigure}

    \caption{An illustration of representing knowledge considering several entities in KG.}
    \label{fig:kg_examples}
\end{figure*}

Figure~\ref{fig:kg_examples} presents example subgraphs extracted from the constructed KG, illustrating how Graph RAG represents SME-related cybersecurity knowledge through interconnected entities and relationships. Figure 5(a) depicts cybersecurity threat relationships, where the central node Cyber Threat is connected to various attack vectors and misinformation-related risks, including phishing, ransomware, email spoofing, data breaches, social engineering, AI-generated misinformation, and LLM hallucinations. This structure demonstrates how diverse cybersecurity threats and misinformation sources can be semantically linked within the knowledge graph. Figure 5(b) illustrates mitigation-oriented knowledge representation. The central node Mitigation Strategies is linked to key cybersecurity practices recommended for SMEs, including security policies, employee awareness training, access control and multi-factor authentication (MFA), regular software updates and patching, and backup and recovery mechanisms. These relationships enable the Graph RAG framework to retrieve structured evidence regarding cybersecurity best practices and organizational resilience measures. Figure 5(c) represents the potential risks and impacts associated with cybersecurity incidents and misinformation. The node Risks/Impacts is connected to outcomes such as financial loss, reputational damage, operational disruption, loss of customer trust, legal issues, and broader business consequences. This subgraph highlights how the knowledge graph captures cause-and-effect relationships, allowing the retrieval process to reason over the potential consequences of cybersecurity failures within SME environments.

Collectively, these example subgraphs demonstrate how Graph RAG organizes domain knowledge into interconnected semantic structures, facilitating relationship-aware retrieval and enhancing the generation of contextually grounded, explainable, and trustworthy responses for SME cybersecurity decision support.

\subsection{Response Quality and Human Interpretation}
Table \ref{pipe evaluation} presents responses generated by base LLMs and their RAG-enhanced counterparts for SME-related queries. The results demonstrate that while the base LLMs generally provide broad and generic answers based on their pre-trained knowledge, the RAG-based models generate more context-aware, detailed, and domain-specific responses by incorporating information retrieved from external SME knowledge sources. For example, in Q1, the base model offers a general explanation of customer feedback usage, whereas the RAG model introduces structured professional practices derived from SME guidance. Similarly, in Q2, the RAG-enhanced response expands upon the base answer by including additional elements such as empathy, confidentiality, and professional conduct. Most notably, in Q3, the base model provides an incorrect or outdated cyber incident reporting contact, while the RAG model retrieves the correct Australian Cyber Security Centre (ACSC) hotline number from authoritative sources. Overall, the comparison highlights that RAG significantly improves response quality by enhancing completeness, contextual relevance, and factual accuracy, thereby reducing misinformation and increasing the reliability and trust of LLM outputs for SME-specific applications. Trust is built when the model's outputs are not only accurate but also clearly derived from reliable and contextually relevant sources \cite{c25}. By incorporating external, authoritative data, RAG based models can provide more precise, actionable, and up-to-date recommendations, which strengthens user confidence.

In addition, Table \ref{tab:rag-scenarios} illustrates how Graph RAG and Vector RAG retrieve complementary information to answer SME-related cybersecurity queries. Graph RAG extracts structured knowledge in the form of entity–relationship triples, providing explicit semantic connections between concepts, while Vector RAG retrieves relevant contextual passages from the source documents. The examples show that combining both retrieval mechanisms enables the system to generate accurate and comprehensive responses, where Graph RAG contributes factual relationships and Vector RAG provides supporting contextual details. This demonstrates the effectiveness of integrating structured and unstructured knowledge retrieval for improving answer quality in SME-specific domains.

\begin{table*}[htbp!]
\centering
\tiny
\caption{Base LLM and RAG-based LLM Responses for SME-related Queries}
\label{pipe evaluation}
\small

\begin{tabularx}{\textwidth}{@{}p{4cm} X@{}}
\hline
\multicolumn{2}{c}{\textbf{Prompt and Model Responses}} \\
\hline

\multicolumn{2}{l}{\textbf{Q1: How can businesses use customer feedback to improve their services?}} \\
\multicolumn{2}{p{16cm}}{A1 (Mistral-7B): By actively listening, analyzing, and implementing customer feedback to enhance products, services, and customer experience.} \\
\multicolumn{2}{l}{A1 (Mistral-RAG): Focuses on professionalism in client interactions including confidentiality and accurate communication.} \\
\multicolumn{2}{p{16cm}}{Human Interpretation: Base model gives generic feedback usage, while RAG introduces structured strategies from external SME cybersecurity guidance.} \\

\multicolumn{2}{l}{\textbf{Q2: What are the best practices for maintaining professionalism in client interactions?}} \\
\multicolumn{2}{l}{A2 (LLaMA): Communicate clearly, actively listen, and show respect.} \\
\multicolumn{2}{l}{A2 (LLaMA-RAG): Includes structured practices such as active listening, empathy, confidentiality, and professional behavior.} \\
\multicolumn{2}{l}{Human Interpretation: RAG improves completeness by incorporating formal SME communication guidelines.} \\

\multicolumn{2}{l}{\textbf{Q3: In Australia, urgent oral reports of cyber incidents can be made to which number?}} \\
\multicolumn{2}{l}{A3 (Qwen): 1800 CYBER REPORT.} \\
\multicolumn{2}{l}{A3 (Qwen-RAG): 1300 292 371 (ACSC official cyber hotline).} \\
\multicolumn{2}{l}{Human Interpretation: RAG corrects outdated or incorrect information using authoritative SME cybersecurity sources.} \\

\hline
\end{tabularx}
\end{table*}


\begin{table*}[htbp!]
\centering
\tiny
\caption{Examples demonstrating Graph RAG and Vector RAG retrieval scenarios under the SME domain.}
\label{tab:rag-scenarios}
\begin{tabular}{@{}p{4cm} p{10cm}@{}}
\\

\multicolumn{2}{l}{\textbf{Q1: What are the three basic security measures the guide recommends?}} \\

\multicolumn{2}{l}{\textbf{Graph Retrieval:}} \\
\multicolumn{2}{l}{(Multi-factor authentication, is recommended, for Small Businesses)} \\
\multicolumn{2}{l}{(Regular software updates, is recommended, for Small Businesses)} \\
\multicolumn{2}{l}{(Data backup, is recommended, for Small Businesses)} \\

\multicolumn{2}{l}{\textbf{Vector Retrieval:}} \\
\multicolumn{2}{l}{``Small businesses should enable multi-factor authentication to improve security.''} \\
\multicolumn{2}{l}{``Regular software updates help protect systems from vulnerabilities.''} \\
\multicolumn{2}{l}{``Backing up important business data reduces the risk of permanent data loss.''} \\

\multicolumn{2}{l}{\textbf{A1:} The guide recommends turning on multi-factor authentication,} \\
\multicolumn{2}{l}{updating software regularly, and backing up important data.} \\

\\

\multicolumn{2}{l}{\textbf{Q2: Which Australian government agency provides advice and has a 24/7 hotline for cyber security support?}} \\

\multicolumn{2}{l}{\textbf{Graph Retrieval:}} \\
\multicolumn{2}{l}{(ACSC, provides, cyber security advice)} \\
\multicolumn{2}{l}{(ACSC, offers 24/7 hotline, for cyber security support)} \\
\multicolumn{2}{l}{(ACSC, provides, resources for SMEs)} \\

\multicolumn{2}{l}{\textbf{Vector Retrieval:}} \\
\multicolumn{2}{l}{``The Australian Cyber Security Centre (ACSC) provides cyber security guidance''} \\
\multicolumn{2}{l}{``and incident support services for small businesses.''} \\
\multicolumn{2}{l}{``Businesses can contact the ACSC through its 24/7 hotline.''} \\

\multicolumn{2}{l}{\textbf{A2:} The Australian Cyber Security Centre (ACSC) provides advice} \\
\multicolumn{2}{l}{and support, including through a 24/7 hotline.} \\

\\

\multicolumn{2}{l}{\textbf{Q3: What types of scams do cybercriminals use to target small businesses, and what are the goals of these scams?}} \\

\multicolumn{2}{l}{\textbf{Graph Retrieval:}} \\
\multicolumn{2}{l}{(Cybercriminals, use, email scams)} \\
\multicolumn{2}{l}{(Cybercriminals, use, social media scams)} \\
\multicolumn{2}{l}{(Cybercriminals, use, phishing attacks)} \\

\multicolumn{2}{l}{\textbf{Vector Retrieval:}} \\
\multicolumn{2}{l}{``Cybercriminals often impersonate trusted organizations through emails,''} \\
\multicolumn{2}{l}{``phone calls, text messages, and social media platforms.''} \\
\multicolumn{2}{l}{``These scams aim to steal sensitive or financial information from businesses.''} \\

\multicolumn{2}{l}{\textbf{A3:} Cybercriminals use email, text message, phone call, phishing,} \\
\multicolumn{2}{l}{and social media scams to trick small businesses into revealing sensitive} \\
\multicolumn{2}{l}{information or sending money.} \\

\hline
\end{tabular}
\end{table*}


\begin{table*}[t]
\centering
\tiny
\caption{Comparison of Existing Studies with our Work}
\label{tab:lit-summary}
\begin{tabularx}{\textwidth}{@{} p{4cm} X X X X X X }
\toprule
\textbf{Reference} & \textbf{Dataset used} & \textbf{Results} & \textbf{Misinformation Handling} & \textbf{Domain-Specific Knowledge Integration} & \textbf{Hallucination Mitigation} & \textbf{Human Interpretation} \\
\midrule
Han et al. (2024) &
1,025 biomedical facts; USMLE-derived adversarial set &jailbreak success up to 58\%. &
Yes &
Yes &
Yes &
No \\
Wan et al. (2025) &
DfAM dataset &
Exact match accuracy of 77.8\%, context precision of 76.5\% &
No &
Yes &
Yes &
Yes \\

Pendyala \& Hall (2024) &
COVID-19 and LIAR datasets &
80.75\% accuracy on COVID-19; all models perform poorly on Liar Dataset & Yes &
No &
No &
Yes \\

Adel \& Alani (2025) &
124 studies screened, 40 selected  & 
Hallucination rates up to 91\%. & 
No & 
Yes & 
Yes &
Yes \\

Hu et al. (2025) &
~56k generated news articles &
 The ranking of real news declines by up to 31.35\% in the presence of LLM-generated news. &
Yes &
Yes &
No &
Yes\\

Our Paper &
SME PDF &
F1-BertScore 91.52\% with a hallucination rate of 0.28\% &
Yes &
Yes &
Yes&
Yes\\
\bottomrule
\end{tabularx}
\end{table*}

\subsection{Comparative Analysis}
Table~\ref{tab:lit-summary} presents a comparative analysis of recent studies addressing misinformation and hallucination in Large Language Models (LLMs). Prior works have explored hallucination mitigation, domain-specific retrieval, and trustworthy LLM across healthcare, manufacturing, education, and misinformation detection domains. Studies such as Han et al. (2024) and Adel and Alani (2025) highlighted the risks of hallucinations in LLM-generated outputs, while Wan et al. (2025) demonstrated the effectiveness of retrieval-augmented approaches for improving contextual relevance. Similarly, Pendyala and Hall (2024) showed that misinformation detection performance depends heavily on data set characteristics and domain knowledge. In contrast to existing works, our work provides a unified evaluation of misinformation mitigation, domain-specific knowledge integration, hallucination reduction, and trust analysis within the SME cybersecurity domain. The proposed RAG-based framework achieved an F1-BERTScore of 91.52\% with a low hallucination rate of 0.28\%, demonstrating its effectiveness for reliable SME-oriented applications.

\section{Discussion}
This study investigates VectorRAG and GraphRAG modeling approaches to mitigate hallucinations
and misinformation risks and evaluate their effectiveness in SME environments. Our experimental
evaluation is conducted across multiple state-of-the-art LLMs, including LLaMA, Mistral, and Qwen, and the results show significant improvements in model performance. The integration of RAG improved model accuracy by grounding responses in verified external knowledge and showed noticeable improvements in F1-BERTScore, indicating more accurate and human-like responses. 

For example, Mistral-7B increased its F1-BERTScore from 82.45\% to 91.52\%, and LLaMA saw a boost from 74.12\% to 89.05\%. These improvements demonstrate how vector RAG helps models produce more relevant and accurate content by referencing trusted sources. However, graph RAG shows a little bit less performance in this case as the dataset consists of limited PDF documents related to SME in Australia. Although the documents share thematic similarity, explicit cross-document entity relationships are limited. Graph RAG relies on explicit entity–relationship structures to support multi-hop reasoning \cite{c26}. However, due to the low relational density across documents, the constructed knowledge graph remained sparse, limiting its retrieval effectiveness. Vector RAG consistently outperformed Graph RAG in terms of answer relevance and completeness, as semantic similarity was sufficient to capture the underlying information needs of the queries. These results indicate that Graph RAG is highly dependent on the presence of rich, interconnected relational structures. For concept-centric and policy-oriented corpora, vector semantic retrieval remains more effective While Graph RAG underperformed in isolation, future work could explore hybrid approaches where graph constraints are applied selectively for entity verification rather than primary retrieval. Overall, this study highlights the potential of RAG enhancing LLMs with context-specific knowledge for mitigating misinformtaion in SMEs.

\section{Conclusion}
This study examined how context-specific knowledge can be incorporated into LLMs through retrieval-enhanced mechanisms to reduce hallucinations and mitigate misinformation risks in SMEs. By leveraging domain-specific SME documents and evaluating vector RAG and graph RAG across multiple LLM architectures, the research provides practical insights into how retrieval mechanisms influence the quality, reliability, and usability of LLM-generated responses in real-world business environments. The findings indicate that vector RAG is particularly well-suited for SME-focused document collections, where the knowledge is predominantly descriptive, policy-oriented, and loosely interconnected. While graph RAG offers structured reasoning and improved interpretability when rich relationships are present, its effectiveness is inherently dependent on the availability of dense and explicit entity–relationship structures. Future research can be extended to explore adaptive or hybrid retrieval frameworks that dynamically balance semantic and structured knowledge to further enhance robustness, explainability, and trust in LLM-assisted decision-making systems for SMEs.

\section*{Contributions}
Writing - original draft preparation: M.S.I and I.H.S.; critical analysis and design: I.H.S., A.M., H.J.; Experiments: M.S.I; writing-review and editing: C.I., A.M., A.I., I.H.S., and H.J.; project administration and supervision: I.H.S. ; All authors reviewed the manuscript.

\section*{Acknowledgment}
This work is supported by funding from the School of Science, Edith Cowan University, Australia. The authors also acknowledge the use of AI tool for proofreading and improving the language of this manuscript.

\bibliographystyle{model1-num-names}

\bibliography{cas-refs}


\end{document}